\documentclass{SCIS2026}
\usepackage{hyperref}
\usepackage{algorithm}
\usepackage{algorithmic}
\usepackage{float}
\usepackage{CJKutf8}
\floatstyle{ruled}
\restylefloat{algorithm}
\newcounter{algorithmicline}
\renewenvironment{algorithmic}[1][0]{%
  \setcounter{algorithmicline}{0}%
  \begin{list}{\arabic{algorithmicline}:}{%
    \usecounter{algorithmicline}%
    \leftmargin=2.5em%
    \labelwidth=2em%
    \labelsep=0.5em%
    \itemsep=0pt%
    \parsep=0pt%
  }%
}{\end{list}}
\providecommand{\STATE}{}
\providecommand{\FOR}[1]{}
\providecommand{\ENDFOR}{}

\renewcommand{\STATE}{\item}
\renewcommand{\FOR}[1]{\item {\bfseries for} #1 {\bfseries do}}
\renewcommand{\ENDFOR}{\item {\bfseries end for}}

\begin{document}

%%
% Authors do not modify the information below
% 作者不需要修改此处信息
\ArticleType{RESEARCH PAPER}
%%

% title: 标题
\title{SpatioTemporal Causal Network Diagnostics for Geographic Tipping Point Early Warning}{ST-CND}

% Corresponding author: 通信作者
\author[1,\dag,*]{Zhaoyuan Yu}{{yuzhaoyuan@njnu.edu.cn}}
\author[2,\dag]{Zhangyong Liang}{}

% Address. 地址
\address[1]{Jiangsu Center for Collaborative Innovation in Geographical Information Resource Development and Application, Nanjing 210023, China}
\address[2]{National Center for Applied Mathematics, Tianjin University, Tianjin 300072, China}
\contributions{Zhaoyuan Yu and Zhangyong Liang contributed equally to this work.}

% Abstract. 摘要
%\abstract{Geographic tipping points pose severe challenges for localized early warning. Classical spatial indicators (e.g., Moran's $I$) summarize global structure. However, spatial dilution, Euclidean assumptions and correlated noise limit their diagnostic value. This paper introduces SpatioTemporal Causal Network Diagnostics (ST-CND) for tipping-point detection and localization. ST-CND reconstructs directed information-flow topologies via transfer entropy. It estimates local recovery-rate decay using subgraph-constrained dynamic mode decomposition. It also identifies critical subnetworks from elevated fluctuation, synchronization and reduced external coupling. Validated on synthetic and empirical datasets, ST-CND delivers localized and interpretable warnings. Our method performs as well as traditional methods when analyzing Indo‑Pacific sea surface temperatures. When analyzing historical changes in the Atlantic Ocean current (1870–2022), it achieves higher prediction accuracy and better matches the actual regions of change, outperforming traditional methods based on network relationships and autocorrelation. The framework provides an interpretable and scalable pipeline for spatial early warning in Earth system science.}

\abstract{Geographic tipping points in ecosystems, climate subsystems, or ice sheets pose severe challenges for localized early warning. Classical spatial indicators such as Moran's I summarize global spatial structure, but they struggle with three issues: spatial dilution, Euclidean assumptions, and correlated noise. This paper introduces SpatioTemporal Causal Network Diagnostics (ST-CND), a framework that addresses these three issues by representing the geographic field as a time-evolving directed causal network. The core workflow is: (1) infer which spatial nodes help predict other nodes via transfer entropy, replacing fixed Euclidean neighbourhoods with data-driven information-flow topology; (2) estimate local recovery rate within each candidate subnetwork via dynamic mode decomposition; (3) identify the most vulnerable subnetwork by combining three signals — high internal fluctuation, high internal synchronization, and low external coupling — which suppresses false alarms from spatially correlated noise. Validated on synthetic bifurcations and two observational sea-surface temperature benchmarks (Indo-Pacific SST and North Atlantic AMOC), ST-CND delivers localized, interpretable warnings. On the AMOC task, it achieves AUROC 0.783 and critical subnetwork IoU 0.378, outperforming recurrence-network and $\lambda$-AR1 baselines. The framework provides an interpretable and scalable pipeline for spatial early warning in Earth system science.}
  
% Keywords. 关键词

\maketitle

%%
% The main text. 正文部分
%%

\section{Introduction}

%Tipping dynamics are a defining feature of Earth's surface systems.  The Amazon rainforest may cross a deforestation threshold~\cite{Lenton2008,Lovejoy2018}.  Recent observation and modelling studies further link Amazon stability to coupled climate--land-use pressures and Earth-system tipping interactions~\cite{Franco2025AmazonTransformation,Boers2025Destabilization}.  The Sahel may shift to desert under sustained drought pressure~\cite{Scheffer2001}.  The Atlantic meridional overturning circulation (AMOC) may also approach a tipping point with global consequences~\cite{Rahmstorf2005,Boers2021,Ditlevsen2023}.  New AMOC-focused evidence emphasizes structural stability changes, stochastic tipping risk, subpolar North Atlantic destabilization and Gulf Stream precursors~\cite{Dima2025AMOCStructural,Oh2025NoiseInducedAMOC,ArellanoNava2025SubpolarBivalves,vanWesten2026GulfStream}.  These events affect ecosystems, human livelihoods and climate feedbacks~\cite{Lenton2011,Steffen2018,Wunderling2021}.  Early detection therefore matters for scientific understanding, adaptation planning and risk governance.

Tipping dynamics are a defining feature of Earth's surface systems~\cite{Lenton2011,Lovejoy2018}. Well-known examples include the Amazon rainforest crossing a deforestation threshold~\cite{Franco2025AmazonTransformation,Boers2025Destabilization}, the Sahel shifting to desert under drought pressure~\cite{Scheffer2001}, and the Atlantic meridional overturning circulation (AMOC) approaching a tipping point with global consequences~\cite{Dima2025AMOCStructural,Oh2025NoiseInducedAMOC,ArellanoNava2025SubpolarBivalves,vanWesten2026GulfStream}. These events affect ecosystems, human livelihoods, and climate feedbacks~\cite{Wunderling2021}. Early detection therefore matters for scientific understanding, adaptation planning, and risk governance.
  
In complex systems, critical slowing down (CSD) provides a widely used theoretical basis for tipping-point early warning.  As a system approaches a local bifurcation, its dominant recovery rate from perturbation weakens.  Stochastic perturbations then accumulate, producing rising variance, increasing lag-1 autocorrelation, and spectral reddening in the observed signal~\cite{Scheffer2009,Dakos2012}.  These temporal signatures have been validated in empirical ecosystem experiments~\cite{Carpenter2011}.  For spatially extended systems, CSD has been extended to spatial variance, spatial skewness, and Moran's $I$ for measuring spatial autocorrelation~\cite{Moran1950,Dakos2010,Kefi2014}.  Recent work has also stressed that classical CSD indicators can be ambiguous under nonstationary noise, transient forcing, limited signal-to-noise ratio and alternative stability-loss mechanisms, motivating model-aware and nonequilibrium diagnostics~\cite{Rietkerk2025Ambiguity,Kubo2025Predictability,Kooloth2026TimeIrreversibility}.

However, classical spatial indicators face three challenges in heterogeneous geographic fields.  First, a tipping event may be triggered within a small vulnerable region.  The surrounding landscape may remain stable, so spatial averaging can obscure local precursory signals.  This dilution is pronounced in patchy systems such as dryland vegetation~\cite{Rietkerk2004,Kefi2007}.  It also affects permafrost terrain~\cite{Joughin2014} and monsoon boundaries.  Second, classical indicators assume fixed Euclidean neighbourhoods and isotropic short-range interactions.  Real geographic systems often include teleconnections, or remote forcing pathways spanning large distances.  Examples include Pacific climate teleconnections and monsoon--cryosphere coupling over the Tibetan Plateau.  Third, spatially correlated environmental noise can inflate global autocorrelation independently of resilience loss.  Moran's $I$ is sensitive to this confound because it measures raw spatial coherence.  It does not separate internal coupling from external noise.

To address these challenges, we develop the SpatioTemporal Causal Network Diagnostics (ST-CND) framework, where causality is used in the Wiener--Granger predictive sense~\cite{Granger1969}.  It is distinct from Pearl-style interventional causality~\cite{Pearl2009}, as discussed in Section~\ref{sec:te-limits}.  The core idea is to represent the geographic field as a time-evolving directed causal network.  Rolling-window transfer entropy infers predictive dependencies between spatially separated nodes~\cite{Schreiber2000}.  Subgraph-constrained dynamic mode decomposition (Graph-DMD) then estimates decay rates in causal subnetworks~\cite{Tu2014}.  Finally, the dynamic network signal diagnostics (DNSD) criterion identifies the most vulnerable geographic subnetwork.  This critical subnetwork combines elevated internal fluctuation, enhanced internal synchronization and reduced external coupling~\cite{Chen2012}.  Together, these components yield a temporal warning signal and a spatially resolved critical subnetwork estimate.  ST-CND integrates information-theoretic graph inference, dynamical mode decomposition and network signal diagnostic scoring.  This design provides an interpretable early-warning framework from an information science perspective.  The paradigm is not limited to geographic tipping elements.  Spatially extended ecological, epidemiological or social systems could be monitored through the same causal topology.  Such use requires data that support meaningful transfer-entropy graph inference.

\begin{figure}[t]
  \centering
  \includegraphics[width=\textwidth]{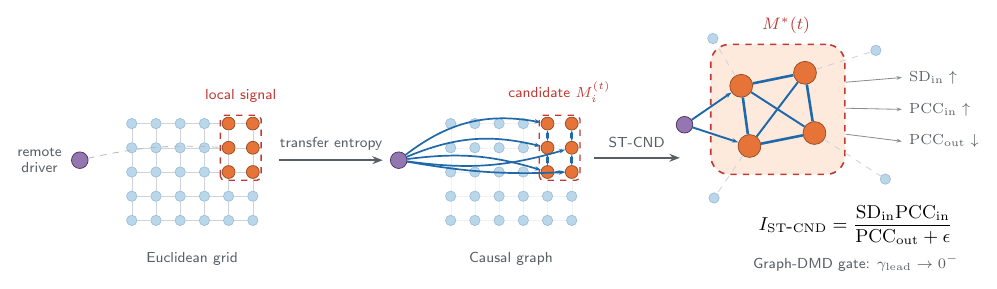}
  \caption{Schematic of ST-CND.  Left: gridded geographic field with a remote driver.  Middle: transfer-entropy information-flow graph.  Right: critical subnetwork~$M^*(t)$ selected by high internal fluctuation, high internal synchronization, weak external coupling and slow Graph-DMD recovery.}
  \label{fig:cgdnb-schematic}
\end{figure}

The contributions of this paper are as follows:
\begin{itemize}
  \item We introduce a directed causal graph for geographic tipping-point warning.  It is inferred by transfer entropy with surrogate-thresholded edges.  This formulation replaces fixed Euclidean neighbourhoods with data-driven teleconnection structures.
  \item We derive a ST-CND risk score that combines internal variance, internal synchronization and external decoupling.  This three-signal design suppresses false alarms induced by spatially correlated red noise, a known limitation of classical spatial indicators.
  \item Graph-DMD and a logistic readout are integrated to connect statistical precursors with a physically interpretable recovery-rate decay and feature-level attribution analysis.
  \item We design a reproducible validation protocol spanning synthetic bifurcations, reaction--diffusion fields and teleconnected systems.  It also includes two observational benchmarks, Indo-Pacific SST and AMOC.  On the AMOC task, ST-CND achieves AUROC 0.783 and critical subnetwork IoU 0.378, outperforming recurrence-network EWS~\cite{Boers2021} and $\lambda$-AR1~\cite{Ditlevsen2023} baselines.
\end{itemize}

The paper is organized as follows: Section~2 reviews related work on spatial early warning signals and causal-topological methods, Section~3 presents the ST-CND methodology, Section~4 reports experimental results and discusses implications and limitations, Section~5 validates and applies ST-CND in real-world spatial resilience contexts, and Section~6 concludes the paper.

\section{Related work}

\subsection{Classical early warning signals and their spatial limitations}

A widely used foundation for tipping-point prediction is CSD.  In this setting, a system's recovery rate approaches zero near a bifurcation.  Pioneering studies identified increased variance, lag-1 autocorrelation and spectral reddening as generic temporal indicators~\cite{Scheffer2009,Dakos2012}.  Spectral reddening and EWS monotonicity are standard pre-processing or summary options for classical temporal EWS, but they are not used here as primary indicators.  For spatially extended systems, these ideas have been generalized to spatial variance, skewness and Moran's $I$~\cite{Moran1950,Dakos2010,Kefi2014}.  Spatial pattern indicators have also been used in dryland ecosystems~\cite{Rietkerk2004,Kefi2007}.  Examples include vegetation patch-size distribution and spatial correlation length.  Remote-sensing studies further show that resilience loss may be spatially heterogeneous~\cite{Boers2021,Boulton2022,Rietkerk2004}.  Recent syntheses and methodological studies show that such signals require careful interpretation because interactions among tipping elements, sensor heterogeneity, seasonality and nonstationary noise may mask or mimic genuine destabilization~\cite{Rietkerk2025Ambiguity,Boers2025Destabilization,Smith2026Instabilities}. 

However, standard spatial EWS rely on fixed spatial weight matrices.  
Examples include 4- or 8-neighbourhood grids (commonly termed rook and queen contiguity, respectively, by analogy with chess-piece moves).  
This constrains their perspective to localized isotropic diffusion.  Consequently, they struggle in heterogeneous environments where tipping can be triggered in a small subregion while the surrounding field remains stable.  In such cases, a global statistic may obscure localized precursory changes.  Spatially correlated red noise can also mimic increasing spatial coherence.  This occurs without true resilience loss and is problematic for global autocorrelation measures.  These limitations motivate the need for localized, topology-aware indicators that can distinguish genuine resilience loss from passive spatial coherence.

\subsection{Causal-topological and interpretable warning methods}

Transfer entropy~\cite{Schreiber2000} captures directional information flow in predictive networks, while climate-network approaches~\cite{Donges2009} map atmospheric teleconnections without relying on physical distance.  Recurrence-network indicators have detected resilience changes in AMOC ~\cite{Boers2021,Ditlevsen2023}, and controlled simulations reveal that CSD indicators can fail under misspecified noise or non-adiabatic forcing~\cite{Ashwin2012,Boettiger2012}.  
Dynamical causal network theory~\cite{Chen2012} identifies pre-transition states via a characteristic subnetwork whose internal correlation rises and external correlation falls.
However, its original biomedical framing assumed $N\sim 50$ tightly co-regulated variables, and extending it to spatial Earth systems where $N$ can exceed several hundred and $|\overline{M}_i|\gg |M_i|$ requires explicit statistical treatment.  
PCMCI provides stronger confounding control than bivariate transfer entropy but its computational cost grows with conditioning-set size, making it difficult to apply over hundreds of spatial nodes in rolling windows~\cite{Runge2019}.
DMD~\cite{Tu2014} and its variants (Hankel-DMD~\cite{Brunton2017HAVOK}, sparse DMD) extract Koopman eigenvalues from spatiotemporal fields, while phase-space reconstruction and recurrence quantification analysis (RQA) offer complementary geometric recurrence indicators~\cite{Marwan2007}.  Recent Floquet/eigenvalue-based work further shows how stability can be estimated directly from seasonal and spatiotemporal data without aggressive detrending~\cite{Smith2026Instabilities}.  Deep-learning classifiers trained on simulated bifurcations show high sensitivity~\cite{Bury2021,zhuge2025deep} but extracted features need not align with physical CSD quantities, limiting process interpretation in geographic risk management.

Motivated by these studies, ST-CND integrates directed information-flow topology, Dynamical causal network scoring and DMD-based recovery-rate estimation into a unified framework.  Unlike existing methods, ST-CND simultaneously addresses spatial localization, teleconnection recovery and red-noise robustness.  Its logistic readout preserves physical interpretability, which is essential for risk communication and adaptive management in geographic applications.

\section{Methodology}
At its core, ST-CND implements three simple ideas. First, the classical assumption that only nearby locations interact is replaced by a data-driven causal graph: transfer entropy reveals which spatial nodes actually help predict each other, regardless of geographic distance. Second, within each discovered neighbourhood, dynamic mode decomposition measures how quickly perturbations decay — a fast decay indicates resilience, while slow decay signals vulnerability. Third, the most at-risk subnetwork is identified by combining three signals: high internal fluctuation, high internal synchronization, and weak coupling to the external field. This three-signal design suppresses false alarms from spatially correlated noise, which is a known weakness of classical spatial indicators. The following subsections formalize each step.

\subsection{Problem formulation}
\label{sec:problem}

We consider a spatially extended dynamical system. Let $\mathbf{s}_t\in\mathbb{R}^N$ denote the latent system state at $N$ spatial nodes at time $t$. The system evolves under a slowly varying control parameter $\mu(t)$:
\begin{equation}
  \mathbf{s}_{t+1} = \mathbf{F}(\mathbf{s}_t; \mu(t)) + \boldsymbol{\eta}_t, \label{eq:dynamics}
\end{equation}
where $\mathbf{F}$ is a differentiable map governing the spatial interactions and $\boldsymbol{\eta}_t$ is stochastic forcing. The control parameter $\mu(t)$ changes on a timescale much longer than the characteristic dynamics, so the system remains locally near a steady state $\mathbf{s}^*(\mu)$ satisfying $\mathbf{F}(\mathbf{s}^*; \mu) = \mathbf{s}^*$ for each fixed $\mu$.

A \emph{tipping point} occurs when $\mu(t)$ crosses a critical value $\mu_c$, causing the steady state $\mathbf{s}^*(\mu)$ to vanish or destabilize via a local bifurcation. Near $\mu_c$, the dominant eigenvalue of $\partial\mathbf{F}/\partial\mathbf{s}$ approaches $1$ from below, which implies that perturbations decay arbitrarily slowly — this is the \emph{critical slowing down} (CSD) phenomenon that forms the theoretical basis for early warning.

We do not observe $\mathbf{s}_t$ directly. Instead, we have a discrete spatiotemporal record $\{\mathbf{x}_t\}_{t=1}^T$, where
\begin{equation}
  \mathbf{x}_t = \mathbf{s}_t + \boldsymbol{\epsilon}_t, \qquad \mathbf{x}_t\in\mathbb{R}^N, \label{eq:observation}
\end{equation}
and $\boldsymbol{\epsilon}_t$ represents measurement noise and unresolved sub-grid processes. The central challenge is: \emph{given only the noisy observations $\{\mathbf{x}_t\}$, detect whether the system is approaching a tipping point, estimate when it will occur, and identify which spatial subregion is most vulnerable.}

Because $\mu(t)$ drifts slowly, the dynamics are approximately stationary within short windows. We adopt a rolling-window approach. For window size $w$ and terminal index $t$, the snapshot matrix is
\begin{equation}
  \mathbf{X}^{(t)} = [\mathbf{x}_{t-w+1},\ldots,\mathbf{x}_t] \in\mathbb{R}^{N\times w}. \label{eq:snapshot}
\end{equation}
Slow trends are removed by Gaussian smoothing $\mathcal{G}_\sigma$ along the time axis:
\begin{equation}
  \tilde{\mathbf{X}}^{(t)} = \mathbf{X}^{(t)} - (\mathcal{G}_\sigma \ast \mathbf{X}^{(t)}), \label{eq:residual}
\end{equation}
where $\ast$ denotes convolution. The residual matrix $\tilde{\mathbf{X}}^{(t)}$ isolates fast fluctuations around the slow drift, suppressing spurious dependencies from shared low-frequency forcing. All subsequent inference uses $\tilde{\mathbf{X}}^{(t)}$.

Within each window, ST-CND solves three subproblems.  First, causal topology reconstruction infers a directed adjacency matrix $\mathbf{A}^{(t)}$ from $\tilde{\mathbf{X}}^{(t)}$ via transfer entropy, then derives candidate causal neighbourhoods $M_i^{(t)} = \{j : A_{ij}^{(t)}+A_{ji}^{(t)}>0\}\cup\{i\}$ — the set of nodes that significantly exchange information with node $i$.  Second, subgraph-constrained physical decay extraction extracts the leading decay rate $\gamma_{\mathrm{lead}}^{(t)}(M_i)$ for each candidate $M_i^{(t)}$ via subgraph-constrained DMD; under CSD, $\gamma_{\mathrm{lead}}^{(t)}\to0^{-}$ signals approaching instability.  Third, critical subnetwork identification and probability fusion combines internal fluctuation, internal synchronization and external decoupling into a DNSD score, selects the maximizer $M^*(t)$, and fuses its five features into a tipping probability $\mathcal{P}_{\mathrm{tip}}(t)$ via a calibrated logistic readout.  The following subsections provide the estimation details for each subproblem.

\subsection{Causal topology reconstruction}
\label{sec:te-limits}
This step turns the spatiotemporal field into a directed graph whose edges indicate predictive information flow. Here we map each rolling data window $\tilde{\mathbf{X}}^{(t)}$ to a directed, time-varying causal graph. The graph captures non-local dependencies beyond Euclidean proximity. We estimate this topology by calculating transfer entropy (TE) on residual time series.  This step prevents slowly varying global trends from dominating predictive relationships. 
For any node pair $j \to i$, TE asks whether the past of source node $j$ improves one-step prediction of target node $i$ beyond what node $i$'s own past already explains.
It is defined as:
\begin{equation}
  T_{j\to i}^{(t)}=\sum_{\Omega} p\left(\tilde{x}_{i,\tau+1},\tilde{\mathbf{x}}_{i,\tau}^{(h)},\tilde{\mathbf{x}}_{j,\tau}^{(h)}\right) \log \frac{p\left(\tilde{x}_{i,\tau+1}\mid\tilde{\mathbf{x}}_{i,\tau}^{(h)},\tilde{\mathbf{x}}_{j,\tau}^{(h)}\right)}{p\left(\tilde{x}_{i,\tau+1}\mid\tilde{\mathbf{x}}_{i,\tau}^{(h)}\right)},
  \label{eq:te}
\end{equation}
where $\tau\in[t-w+1,t-1]$, $\tilde{\mathbf{x}}_{i,\tau}^{(h)}=(\tilde{x}_{i,\tau},\ldots,\tilde{x}_{i,\tau-h+1})$ is the $h$-step residual history of node $i$, $\Omega$ is the joint support sampled within the current window and $\log$ denotes the natural logarithm. In practice, the finite-sample raw estimate $\widehat{T}_{j\to i}^{(t)}$ is inherently positively biased for short windows. To achieve proper statistical calibration, we compute the effective transfer entropy by subtracting the empirical baseline derived from phase-randomized surrogate source series:
\begin{equation}
  \widehat{T}^{\mathrm{eff},(t)}_{j\to i}=\widehat{T}^{(t)}_{j\to i}-\frac{1}{S}\sum_{s=1}^{S}\widehat{T}^{(s,t)}_{j\to i}.
  \label{eq:effective-te}
\end{equation}
Here, $S$ is the number of surrogate realizations and $\widehat{T}^{(s,t)}_{j\to i}$ is recomputed after replacing the source series by its $s$-th surrogate, which preserves simple temporal structure but destroys directed coupling. The resulting surrogate quantiles are converted to edge-level $p$-values.  Let $p_{j\to i}^{(t)}$ be this surrogate-based edge $p$-value and let $q_{\mathrm{BH}}^{(t)}$ be the Benjamini--Hochberg rejection threshold over ordered node pairs in window $t$.  A multiple-testing correction restricts the false discovery rate (FDR).  This yields a statistically calibrated, sparse directed adjacency matrix $\mathbf{A}^{(t)} \in \mathbb{R}^{N\times N}$:
\begin{equation}
  A_{ij}^{(t)}=\widehat{T}^{\mathrm{eff},(t)}_{j\to i}\,\mathbf{1}\!\left(p_{j\to i}^{(t)}\le q_{\mathrm{BH}}^{(t)}\right).
  \label{eq:adjacency}
\end{equation}
Here rows are target nodes and columns are source nodes; thus $A_{ij}^{(t)}$ records information flow from node $j$ to node $i$. Based on significant directed dependencies, we define each node's local causal neighbourhood $M_i^{(t)}$ by symmetric membership:
\begin{equation}
  M_i^{(t)}=\left\{j\in\{1,\ldots,N\}: A_{ij}^{(t)}>0\ \mathrm{or}\ A_{ji}^{(t)}>0\right\}\cup\{i\}.
  \label{eq:causal-nbhd}
\end{equation}
This symmetrization is used only to form candidate subnetworks: node $j$ is included if it either sends information to, or receives information from, node $i$.  The directed edge weights are still preserved for attribution and visualization.  Thus, a directed edge means that the source history improves prediction of the target, not that an intervention on the source would necessarily change the target. 
For large-scale systems, surrogate-based significance testing can face resolution limits, so the module uses a soft-weight protocol that extracts hub subnetworks from the strongest positive transfer entropy values.  In this protocol, all ordered node pairs with positive KSG transfer entropy are ranked within each rolling window, and only the largest $m_{\max}$ directed pairs are retained when the number of positive pairs exceeds the computational cap.  The retained weighted pairs define incident in- and out-neighbourhoods for each node.  No fixed candidate size is imposed; instead, $|M_i^{(t)}|$ is determined by the retained incident pairs of node $i$, after which empty or singleton candidates are discarded and duplicate memberships are merged before scoring.

%Eq.~\eqref{eq:te} measures Wiener--Granger predictive causality~\cite{Granger1969}, not Pearl-style interventional causality~\cite{Pearl2009}.  Hidden common drivers can inflate bivariate transfer entropy (TE); we therefore interpret $\mathbf{A}^{(t)}$ as a directed information-flow graph rather than a structural causal graph.  Conditional or multivariate TE (e.g., PCMCI~\cite{Runge2019}) can mitigate confounding but is left to future work.  When the number of candidate directed pairs $m$ is large, surrogate-based BH discovery is limited by empirical-$p$ resolution.  In such cases we use a reduced soft-weight validation protocol based on positive KSG TE values (or positive effective TE when surrogates are available), and extract hub subnetworks from the strongest weighted in-/out-neighbourhoods.  All large-grid real-data results in this paper use this disclosed soft-weight protocol.

\subsection{Subgraph-constrained physical decay extraction}

To quantify localized resilience loss, we estimate a dominant decay proxy within each candidate subnetwork.  Because global DMD is vulnerable to spatial dilution, we apply subgraph-constrained DMD (Graph-DMD) on each $M_i^{(t)}$.  Graph-DMD fits the best local linear rule that advances the selected subnetwork from one time step to the next; its eigenvalues summarize whether perturbations inside that subnetwork decay, persist or grow.

Let $\tilde{\mathbf{X}}_{M_i}^{(t)}\in\mathbb{R}^{|M_i|\times w}$ be the residual matrix restricted to nodes in $M_i^{(t)}$.  The matrix $\mathbf{Z}_{1,M_i}$ contains the first $w-1$ residual snapshots, whereas $\mathbf{Z}_{2,M_i}$ contains the same window shifted one step forward:
\[
  \mathbf{Z}_{1,M_i}=\tilde{\mathbf{X}}_{M_i,1:w-1}^{(t)},\qquad
  \mathbf{Z}_{2,M_i}=\tilde{\mathbf{X}}_{M_i,2:w}^{(t)}.
\]
We seek a local linear evolution operator $\mathbf{K}_{M_i}$ such that
\begin{equation}
  \mathbf{Z}_{2,M_i}\approx\mathbf{K}_{M_i}\mathbf{Z}_{1,M_i}.
  \label{eq:local-koopman-map}
\end{equation}
This approximation is interpreted as a finite-window local linearization of the subnetwork dynamics.

After truncating to rank $r$, we compute the singular value decomposition $\mathbf{Z}_{1,M_i}\approx\mathbf{U}_{i,r}\mathbf{S}_{i,r}\mathbf{V}_{i,r}^\top$, where $\mathbf{S}_{i,r}$ is the diagonal singular-value matrix.  The low-dimensional DMD operator is
\[
  \tilde{\mathbf{K}}_{M_i}=\mathbf{U}_{i,r}^\top\mathbf{Z}_{2,M_i}\mathbf{V}_{i,r}\mathbf{S}_{i,r}^{-1}.
\]
Let $\lambda_{i,k}$ be the discrete-time eigenvalues of $\tilde{\mathbf{K}}_{M_i}$.  We map them to continuous-time growth/decay rates via
\[
  \omega_{i,k}=\frac{\ln(\lambda_{i,k})}{\Delta t}.
\]
The leading real part provides a dominant decay proxy,
\begin{equation}
  \gamma_{\mathrm{lead}}^{(t)}(M_i)=\max_k\operatorname{Re}(\omega_{i,k}).
  \label{eq:leading-rate}
\end{equation}
Under local critical slowing down, $\gamma_{\mathrm{lead}}^{(t)}(M_i)\to 0^{-}$ as the subnetwork approaches instability.  A negative $\gamma_{\mathrm{lead}}$ means disturbances decay; values closer to zero mean slower recovery and hence weaker resilience.

\paragraph{Numerical specification.}  Three numerical choices control rank selection, conversion from discrete to continuous time and exclusion of unreliable oscillatory estimates.  First, the DMD rank $r$ uses the Gavish--Donoho optimal hard threshold~\cite{GavishDonoho2014}.  This closed-form estimator separates signal from noise singular values using matrix aspect ratio and noise variance.  The rank is capped at $r\le\min(|M_i|,\,w-1)$.  Robustness is checked by sweeping $r$ over an energy-ratio range of $[0.85,0.99]$.  Section~\ref{sec:results} reports the median $\gamma_{\mathrm{lead}}$.  Second, the sampling interval $\Delta t=1$ throughout.  Third, strongly oscillatory leading modes are excluded from the gate because their inferred decay rate is not a clean recovery-rate estimate.  For short windows where $\min(|M_i|,\,w-1)<10$, we use the de-biased total-DMD variant~\cite{Hemati2017}.

\subsection{CSD scope across experiment classes}  
The strict CSD prediction $\gamma_{\mathrm{lead}}^{(t)}\to 0^{-}$ assumes a saddle-node-like bifurcation.  
This assumption applies to the saddle-node and multivariate recovery-rate experiments.  
The Allen--Cahn experiment is a continuous symmetry-breaking transition.  
Its precursor is the slow homogeneous mode around the symmetric state, not a saddle-node decay rate.  
Therefore, we use Allen--Cahn to benchmark spatially localized nucleation.  
This benchmark exposes the spatial-dilution issue motivating ST-CND.  We interpret $\gamma_{\mathrm{lead}}^{(t)}$ as a slow-mode proxy rather than a literal decay rate.  For teleconnected and red-noise experiments, $\gamma_{\mathrm{lead}}^{(t)}$ denotes the leading continuous-time growth/decay-rate proxy inferred from DMD eigenvalues within $M_i^{(t)}$.  In these cases, the saddle-node limit is one special case rather than a universal target.

\subsection{Critical subnetwork identification via ST-CND}

This step scores each candidate neighbourhood and picks the most vulnerable one based on three signals. Here we convert each candidate subnetwork $M_i^{(t)}$ into a scalar risk score. 
The score identifies the most vulnerable component within the system. 
Our design follows the dynamic network signal diagnostics (DNSD) principle. 
Near a critical transition, a vulnerable subnetwork shows increased internal fluctuation and synchronization. 
It also shows decreased coupling to the background environment. 
To operationalize this principle, we first evaluate windowed sample standard deviations $\hat\sigma_t(k)$. 
We also compute absolute Pearson correlation coefficients $|\hat\rho_t(k,l)|$ on $\tilde{\mathbf{X}}^{(t)}$. 
Absolute correlations allow anti-phase synchronization to contribute to coherence. 
This is useful for dipole modes that indicate shared dynamical constraints. 
For each candidate $M_i^{(t)}$, we also define its complement $\overline{M}_i^{(t)}$ and compute three statistics: average internal standard deviation, average absolute internal correlation and average absolute external coupling. 
Specifically, $\hat\sigma_t(k)$ is the sample standard deviation of node $k$ within window $t$, and $\hat\rho_t(k,l)$ is the Pearson correlation between nodes $k$ and $l$ in the same residual window.  The three DNSD components are defined as
\begin{equation}
  \begin{aligned}
  \mathrm{SD}_{\mathrm{in}}^{(t)}(M_i)
  &=\frac{1}{|M_i|}\sum_{k\in M_i}\hat\sigma_t(k),\\
  \mathrm{PCC}_{\mathrm{in}}^{(t)}(M_i)
  &=\frac{1}{|M_i|(|M_i|-1)}
    \sum_{\substack{k,l\in M_i\\k\ne l}}\left|\hat\rho_t(k,l)\right|,\\
  \mathrm{PCC}_{\mathrm{out}}^{(t)}(M_i)
  &=\frac{1}{|M_i|\,|\overline{M}_i|}
    \sum_{\substack{k\in M_i\\l\in\overline{M}_i}}\left|\hat\rho_t(k,l)\right|.
  \end{aligned}
  \label{eq:dnsd-components}
\end{equation}

These three statistics are combined so that the score increases when a candidate module fluctuates strongly and coherently, but decreases when the same coherence is also present in the background:
\begin{equation}
  I_{\mathrm{ST-CND}}^{(t)}(M_i)= \frac{\mathrm{SD}_{\mathrm{in}}^{(t)}(M_i)\cdot \mathrm{PCC}_{\mathrm{in}}^{(t)}(M_i)}{\mathrm{PCC}_{\mathrm{out}}^{(t)}(M_i)+\epsilon},
  \label{eq:dnb-score}
\end{equation}
where $\epsilon>0$ is a small stabilization constant. This ratio acts as a common-mode rejection mechanism.  Under spatially homogeneous red noise, external correlation inflates internal and background coherence together.  Eq.~\eqref{eq:dnb-score} suppresses such non-specific inflation and isolates localized vulnerable modules. 
To incorporate temporal relaxation, we couple DNSD evidence with $\gamma_{\mathrm{lead}}^{(t)}(M_i)$.  Because critical slowing down makes $|\gamma_{\mathrm{lead}}|$ small, the gate boosts subnetworks that are both internally noisy/coherent and slow to recover.  We apply this gate only to stable-side estimates, collected in $\mathcal{C}_t^-=\{M_i:\gamma_{\mathrm{lead}}^{(t)}(M_i)<0\}$.  For $M_i\in\mathcal{C}_t^-$, the decay-gated risk metric is
\begin{equation}
  R_{\mathrm{ST-CND}}^{(t)}(M_i)= \frac{I_{\mathrm{ST-CND}}^{(t)}(M_i)}{-\gamma_{\mathrm{lead}}^{(t)}(M_i)+\eta},
  \label{eq:decay-gated-score}
\end{equation}
where $\eta>0$ prevents divergence. Windows with positive leading rates are excluded before computing the gated score because they do not represent stable-side recovery estimates. 
Finally, the critical subnetwork $M^*(t)$ is selected at each window $t$ by maximizing this risk profile:
\begin{equation}
  M^*(t)=\arg\max_{M_i\in\mathcal{C}_t^-}R_{\mathrm{ST-CND}}^{(t)}(M_i).
  \label{eq:critical-subnetwork}
\end{equation}

\subsection{Spatiotemporal probability fusion and interpretable readout}

Finally, we fuse the topological risk and dominant decay rate of the critical subnetwork $M^*(t)$.  All five entries below are computed for the selected critical subnetwork in the same rolling window:
\[
  \mathbf{f}_t=
  \left[
  \gamma_{\mathrm{lead}}^{(t)}(M^*(t)),
  I_{\mathrm{ST-CND}}^{(t)}(M^*(t)),
  \mathrm{SD}_{\mathrm{in}}^{(t)}(M^*(t)),
  \mathrm{PCC}_{\mathrm{in}}^{(t)}(M^*(t)),
  \mathrm{PCC}_{\mathrm{out}}^{(t)}(M^*(t))
  \right]^\top,
\]
A regularized logistic classifier maps the operational ST-CND feature vector $\mathbf{f}_t$ to a continuous spatiotemporal tipping probability $\mathcal{P}_{\mathrm{tip}}\in[0,1]$.  This is a calibrated S-shaped conversion from five interpretable indicators to a warning probability:
\begin{equation}
  \mathcal{P}_{\mathrm{tip}}(t)=
  \frac{1}{1+\exp\left[-\left(\boldsymbol{\beta}^\top\mathbf{f}_t+\beta_0\right)\right]}.
  \label{eq:probability}
\end{equation}
Before fitting, the five features are standardized, so the signs and relative magnitudes of $\boldsymbol{\beta}$ provide feature-level attribution for this five-feature ST-CND readout.  This explicit modeling makes the warning mechanism physically transparent.  It complements higher-capacity but less inspectable deep-learning representations~\cite{Bury2021,zhuge2025deep}.

For interpretability analysis only, we also train an auxiliary logistic readout on an expanded diagnostic library that includes conventional EWS statistics, recurrence-quantification indicators and DMD-derived quantities.  This auxiliary model is used to compare feature families in Fig.~\ref{fig:ml-weights}; it is not the operational ST-CND probability model defined above.

% \paragraph{Training and validation protocol.}  
Coefficients $(\boldsymbol{\beta},\beta_0)$ are estimated on 1{,}000 stochastic saddle-node realizations only, with pre-tipping windows ($|\mu(t)-\mu_c|<0.1$) positive and post-tipping windows excluded; here $\mu_c$ is the critical value of the normalized control parameter.  All other experiments use these fixed operational weights.  The classifier uses elastic-net mixing parameter $\rho_{\mathrm{EN}}=0.5$ with penalty strength $\lambda_{\mathrm{EN}}$ selected by 10-fold blocked cross-validation.  Confidence intervals are 95\% bootstrap percentiles ($B=1{,}000$).  Variance inflation factors confirm acceptable multicollinearity (VIF $\le 10$).

% \paragraph{Alarm-threshold and lead-time rule.}  
For Table~\ref{tab:benchmark}, alarm threshold $\tau_{\mathrm{alarm}}$ is the $1-\alpha_{\mathrm{alarm}}$ quantile of $\mathcal{P}_{\mathrm{tip}}$ on no-tipping red-noise fields, with $\alpha_{\mathrm{alarm}}=0.05$.  An alarm starts at the first time $t^\star$ with $\kappa=3$ consecutive threshold crossings, while isolated crossings are ignored.  Lead time is $t_c-t^\star$ in rolling windows.  Here, $t_c$ denotes tipping time, and lead time is zero if $t^\star\ge t_c$.

\subsection{Algorithmic summary}

The complete ST-CND pipeline is summarized in Algorithm~\ref{alg:cgdnb}.
In each rolling window, ST-CND turns the residual field into candidate causal neighbourhoods.  It then evaluates each neighbourhood with DMD and DNSD, selects the highest-risk subnetwork, and converts its features into a calibrated tipping probability.

\begin{algorithm}[t]
\caption{ST-CND: SpatioTemporal Causal Network Diagnostics}
\label{alg:cgdnb}
\begin{algorithmic}[1]
    \STATE {\bfseries Input:} Spatiotemporal field $\mathbf{X}\in\mathbb{R}^{N\times T}$, window size $w$, smoothing scale $\sigma$, IAAFT surrogates $S$, FDR level $\alpha$, DMD rank $r$, retained-pair cap $m_{\max}$, stabilizers $\epsilon,\eta$, fixed readout coefficients $(\boldsymbol{\beta},\beta_0)$
    \STATE {\bfseries Output:} Tipping probability $\mathcal{P}_{\mathrm{tip}}(t)$ and critical subnetwork $M^*(t)$ for each window $t$
    \FOR{each rolling window $t=w,w+1,\ldots,T$}
        \STATE {\bfseries // Spatiotemporal detrending}
        \STATE Detrend: $\tilde{\mathbf{X}}^{(t)}\leftarrow\mathbf{X}^{(t)}-\mathcal{G}_\sigma*\mathbf{X}^{(t)}$
        \STATE {\bfseries // Causal topology reconstruction}
        \STATE Estimate pairwise transfer entropy $\widehat{T}_{j\to i}^{(t)}$ on $\tilde{\mathbf{X}}^{(t)}$
        \STATE If $S>0$, keep BH-significant effective-TE edges in $\mathbf{A}^{(t)}$ and form ego-net candidates $\{M_i^{(t)}\}$
        \STATE If $S=0$, form weighted neighbourhood candidates from the top $m_{\max}$ positive TE values
        \STATE {\bfseries // Subnetwork scoring}
        \FOR{each candidate subnetwork $M_i$}
            \STATE Estimate local recovery rate $\gamma_{\mathrm{lead}}^{(t)}(M_i)$ by Graph-DMD
            \STATE Compute DNSD components $\mathrm{SD}_{\mathrm{in}}^{(t)}$, $\mathrm{PCC}_{\mathrm{in}}^{(t)}$, $\mathrm{PCC}_{\mathrm{out}}^{(t)}$
            \STATE Compute $I_{\mathrm{ST-CND}}^{(t)}(M_i)$ and, if $\gamma_{\mathrm{lead}}^{(t)}(M_i)<0$, the gated risk $R_{\mathrm{ST-CND}}^{(t)}(M_i)$
        \ENDFOR
        \STATE {\bfseries // Critical subnetwork identification and probability fusion}
        \STATE Identify critical subnetwork: $M^*(t)\leftarrow\arg\max_{M_i\in\mathcal{C}_t^-}R_{\mathrm{ST-CND}}^{(t)}(M_i)$
        \STATE Assemble feature vector $\mathbf{f}_t$ and compute $\mathcal{P}_{\mathrm{tip}}(t)$
    \ENDFOR
    \STATE {\bfseries Return} $\{M^*(t),\mathcal{P}_{\mathrm{tip}}(t)\}_{t=w}^{T}$
\end{algorithmic}
\end{algorithm}

\section{Experimental results and discussion}
\label{sec:results}

We design four experiment classes to evaluate ST-CND.  They assess spatial localization, teleconnection recovery, red-noise robustness, physical decay extraction and interpretability.  Additional parameter settings are recorded in the experiment configuration files that accompany the submission package.

\subsection{Experimental settings}

We use four classes of controlled experiments: first, Allen--Cahn reaction--diffusion fields study spatially localized nucleation.  These abstract pattern-forming mechanisms are relevant to vegetation patches and spatial regime shifts~\cite{Rietkerk2004,Kefi2007,Dakos2010,Kefi2014}.  Second, teleconnected spatial systems test whether causal neighbourhoods recover non-local drivers.  Third, spatially correlated random fields without bifurcation serve as no-tipping controls, testing red-noise false alarms.  Fourth, stochastic saddle-node trajectories and controlled multivariate recovery-rate systems evaluate DMD, RQA and ML readouts~\cite{Marwan2007,Tu2014}.

Unless stated otherwise, experiments use the KSG ($k$-NN, $k=4$) TE estimator with history embedding $h=1$.  We use $S=200$ IAAFT surrogates per pair and BH-FDR $\alpha=0.05$.  The histogram estimator is retained only as a robustness option and is not used for the reported observational rows.  The climate-network baseline~\cite{Donges2009} is an undirected Pearson-correlation graph.  Edges connect node pairs in the top $5\%$ absolute correlation, and localization is scored on the largest connected component.

\subsection{Spatiotemporal early warning of tipping points}

This section evaluates ST-CND across temporal, spatial and statistical dimensions of early warning.  The three experiments demonstrate ST-CND's ability to capture localized precursors, recover non-local teleconnections, and control false alarms under red noise.

\begin{figure}[t]
  \centering
  \includegraphics[width=0.75\linewidth]{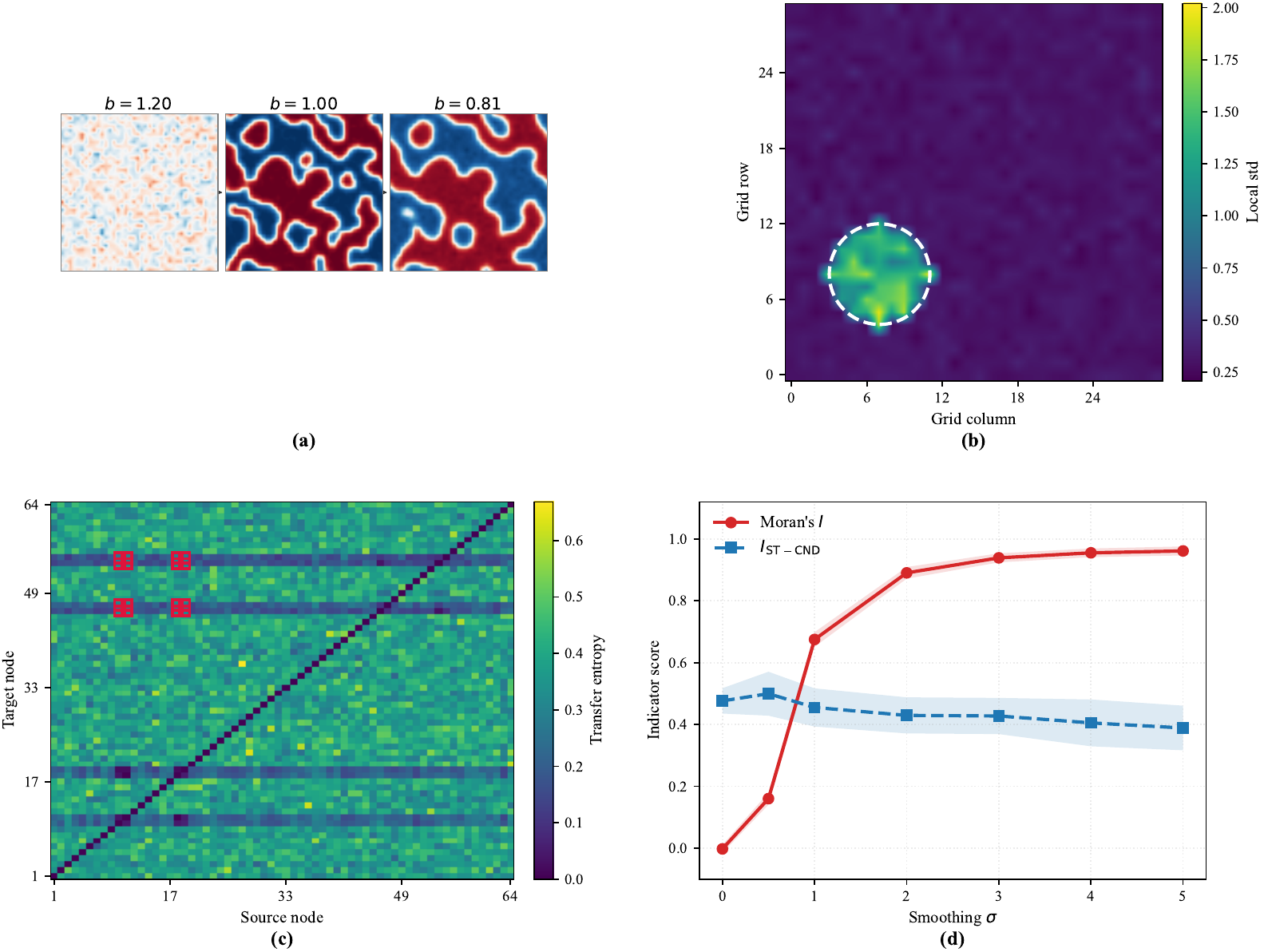}
  \caption{Combined spatial diagnostics: (a) Allen--Cahn evolution from $b=1.20$ to $b=0.80$ with localized nucleation, (b) local standard deviation, (c) transfer-entropy topology, and (d) Moran's $I$ false positives under red noise versus ST-CND response.}
  \label{fig:combined-spatial-properties}
\end{figure}

Figure~\ref{fig:combined-spatial-properties} provides a comprehensive overview of the three evaluation dimensions.  Panel (a) demonstrates the temporal dimension through Allen--Cahn field evolution, showing localized nucleation before macroscopic separation.  Panel (b) presents the spatial dimension via local standard deviation heatmap, with the white dashed circle marking the true nucleation zone.  Panel (c) reveals the causal topology recovered by transfer entropy, where red squares indicate true remote links.  Panel (d) illustrates the statistical dimension, comparing Moran's $I$ false-positive rise under red noise against ST-CND's controlled response.

\subsection{Red-noise robustness of spatial indicators}

\begin{figure}[t]
  \centering
  \includegraphics[width=\linewidth]{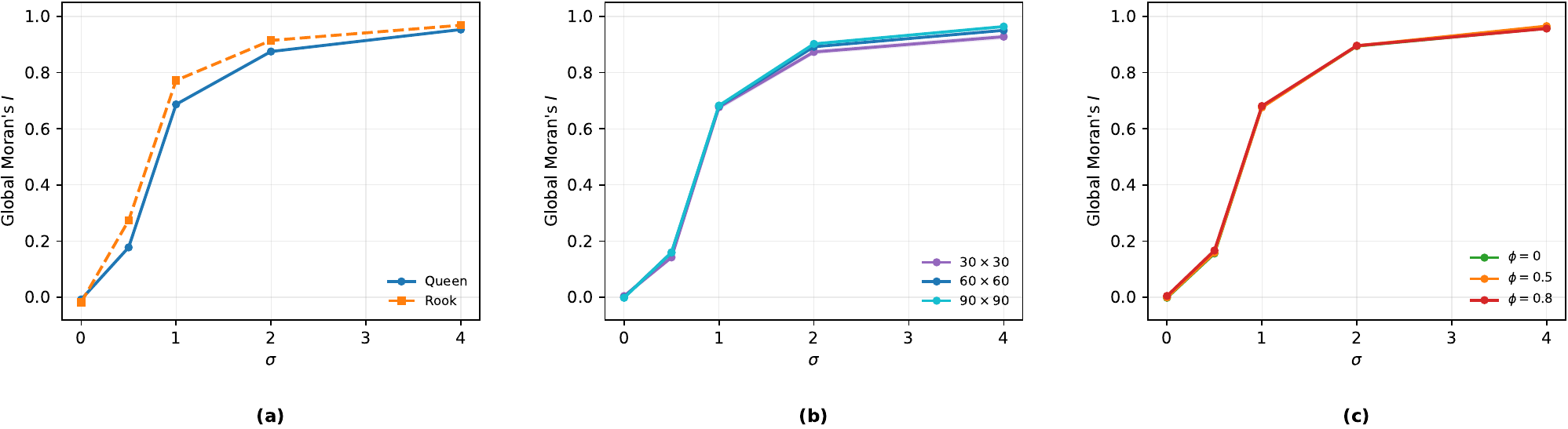}
  \caption{Moran's $I$ on stable fields under Gaussian smoothing: (a) queen versus rook neighbourhoods, (b) grid-resolution sensitivity, and (c) added temporal AR(1) persistence.  Horizontal axes denote smoothing strength $\sigma$.}
  \label{fig:spatial-moran}
\end{figure}

\begin{figure}[t]
  \centering
  \includegraphics[width=\linewidth]{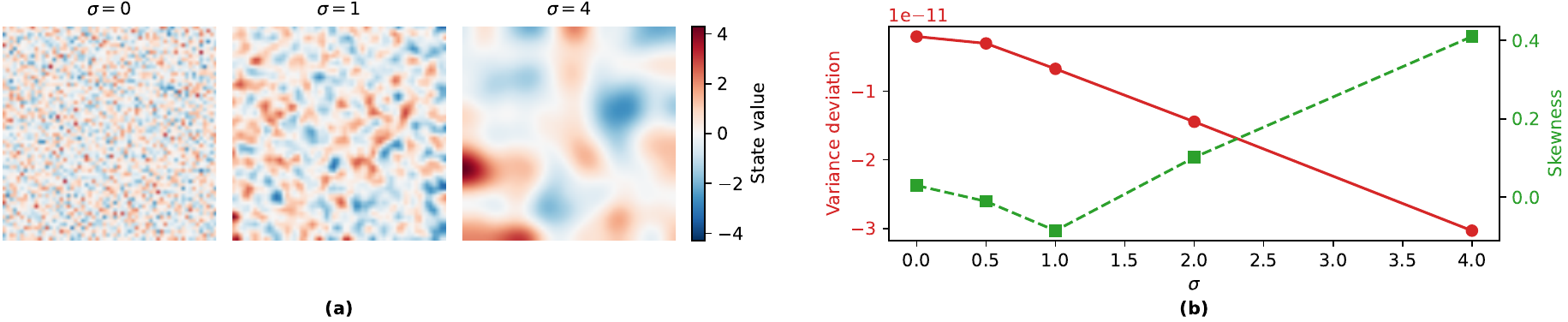}
  \caption{Spatial demo diagnostics for red-noise controls.  (a) shows correlated fields for $\sigma=0$, $\sigma=1$ and $\sigma=4$.  Smoothing alone creates coherent spatial patches without a tipping transition.  (b) reports variance deviation and spatial skewness.}
  \label{fig:spatial-moran-extra}
\end{figure}

\begin{figure}[t]
  \centering
  \includegraphics[width=.9\linewidth]{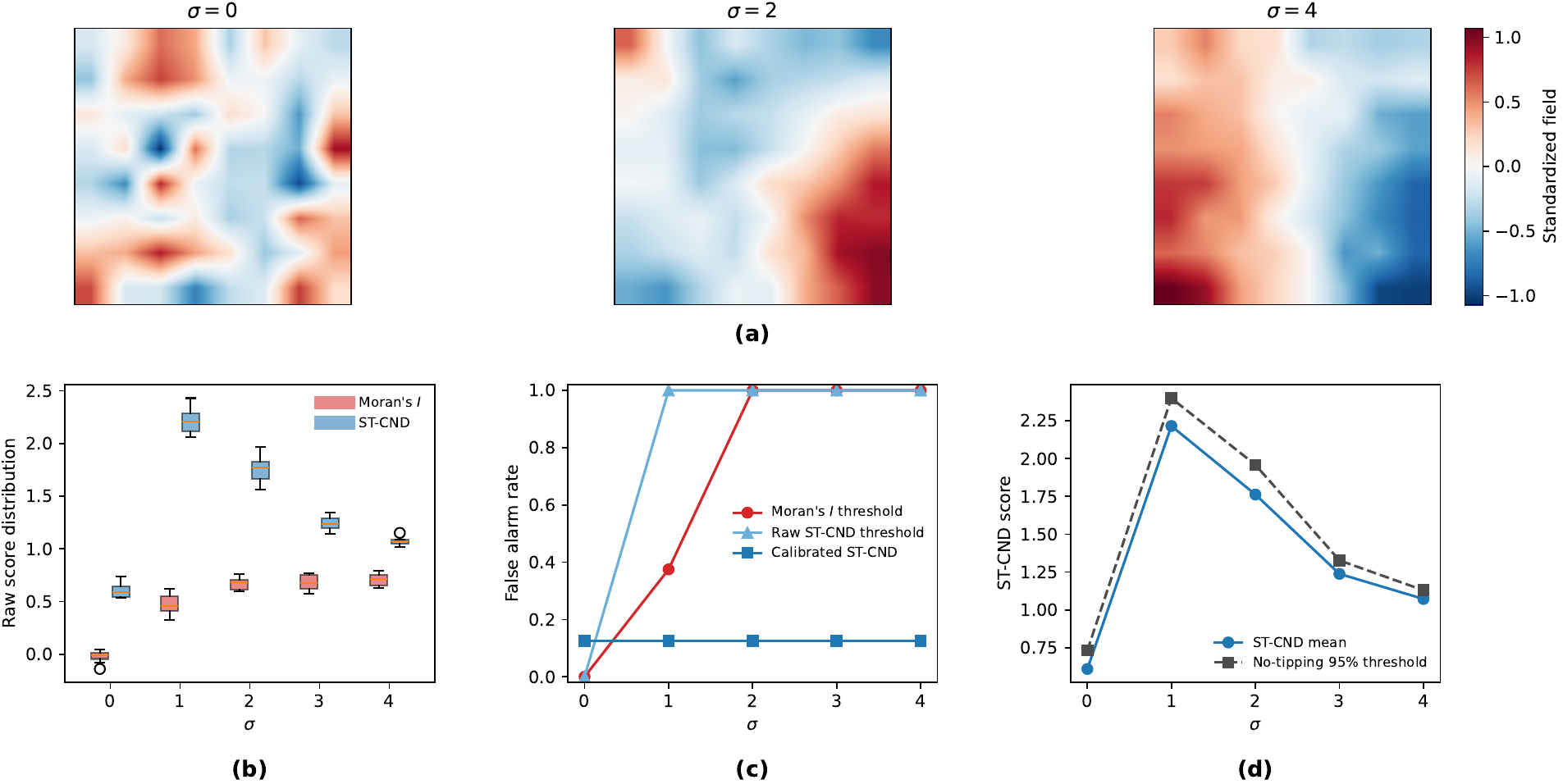}
  \caption{Null calibration under spatially correlated red noise: (a) smoothing with increasing Gaussian strength $\sigma$, (b) raw Moran's $I$ and ST-CND shifts, (c) false alarms from fixed or raw thresholds and their calibrated control, and (d) ST-CND mean below the no-tipping 95\% threshold.}
  \label{fig:cgdnb-null-calibration}
\end{figure}

The spatial Moran's $I$ indicator is sensitive to spatially correlated noise.  To demonstrate this limitation, stable random fields are generated under increasing Gaussian smoothing strength $\sigma$.  Figure~\ref{fig:spatial-moran} shows that Moran's $I$ rises sharply under smoothing.  It increases from near-zero at $\sigma=0$ to approximately 0.95 at $\sigma=4.0$.  This increase is caused by imposed spatial autocorrelation rather than resilience loss.  Figure~\ref{fig:spatial-moran-extra} provides additional spatial diagnostics showing that smoothing alone creates coherent spatial patches without any underlying critical transition.  Figure~\ref{fig:cgdnb-null-calibration} further evaluates no-tipping null calibration, after which the ST-CND alarm rate remains controlled across $\sigma$.  In contrast, raw Moran's $I$ triggers false alarms.

\begin{figure}[t]
  \centering
  \includegraphics[width=\linewidth]{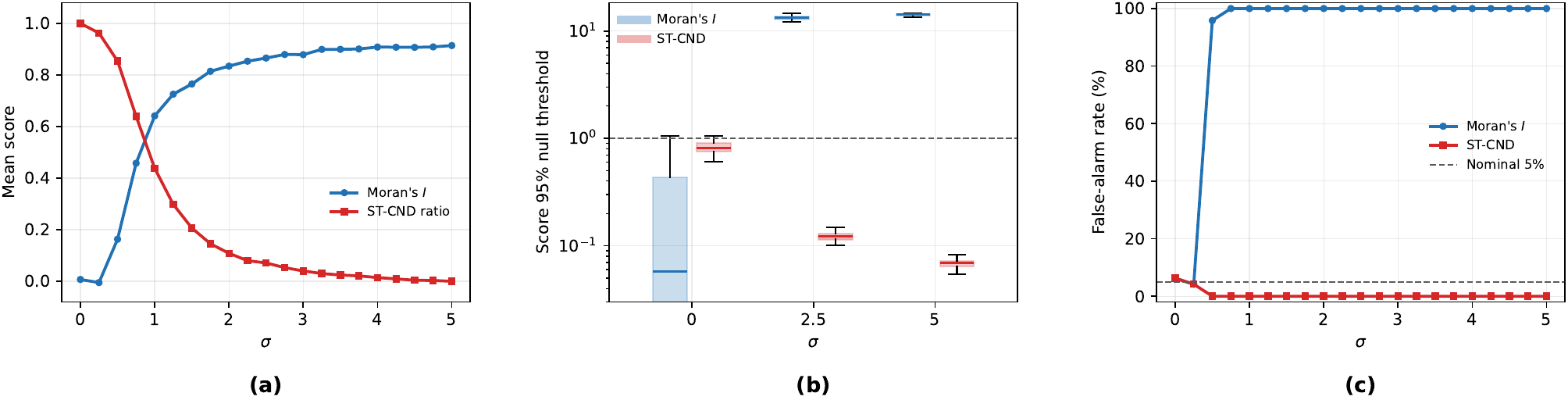}
  \caption{ST-CND versus Moran's $I$ under homogeneous red-noise controls: (a) Moran's $I$ rise and ST-CND ratio decline with increasing smoothing strength $\sigma$, (b) scores normalized by the no-tipping 95\% threshold, and (c) false-alarm rates relative to the nominal 5\% level.}
  \label{fig:dnb-vs-moran-rednoise}
\end{figure}

Figure~\ref{fig:dnb-vs-moran-rednoise} isolates the ratio effect behind this behaviour.  Moran's $I$ increases monotonically with the smoothing scale because the null fields become more spatially coherent.  In contrast, the ST-CND ratio decreases as internal and external correlations rise together under homogeneous red-noise forcing.  Relative to the no-tipping 95\% threshold, Moran's $I$ rapidly produces near-universal false alarms, whereas ST-CND remains near the nominal false-alarm level.  This result supports the use of the external-coupling denominator in Eq.~\eqref{eq:dnb-score} as a guard against spatial-coherence-only alarms.

Table~\ref{tab:moran} quantifies Moran's $I$ across smoothing levels for queen and rook spatial weights.  Queen weights include all eight surrounding cells, whereas rook weights use only the four orthogonal neighbours.  The systematic rise from $-0.0078$ at $\sigma=0$ to $0.9532$ at $\sigma=4.0$ demonstrates a false-positive vulnerability affecting global spatial autocorrelation under smooth red-noise fields.

\begin{table}[H]
\centering
\caption{Moran's $I$ values for stable random fields under increasing smoothing strength.}
\label{tab:moran}
\begin{tabular}{rrrr}
\toprule
Smoothing $\sigma$ & Moran's $I$ (queen) & Moran's $I$ (rook) & Spatial skewness \\
\midrule
0.0 & $-0.0078$ & $-0.0181$ & 0.0294 \\
0.5 & 0.1776 & 0.2734 & $-0.0110$ \\
1.0 & 0.6868 & 0.7721 & $-0.0859$ \\
2.0 & 0.8750 & 0.9144 & 0.1013 \\
4.0 & 0.9532 & 0.9685 & 0.4100 \\
\bottomrule
\end{tabular}
\end{table}

\subsection{Dynamical decay extracted by DMD and RQA}

\begin{figure}[t]
  \centering
  \includegraphics[width=0.95\linewidth]{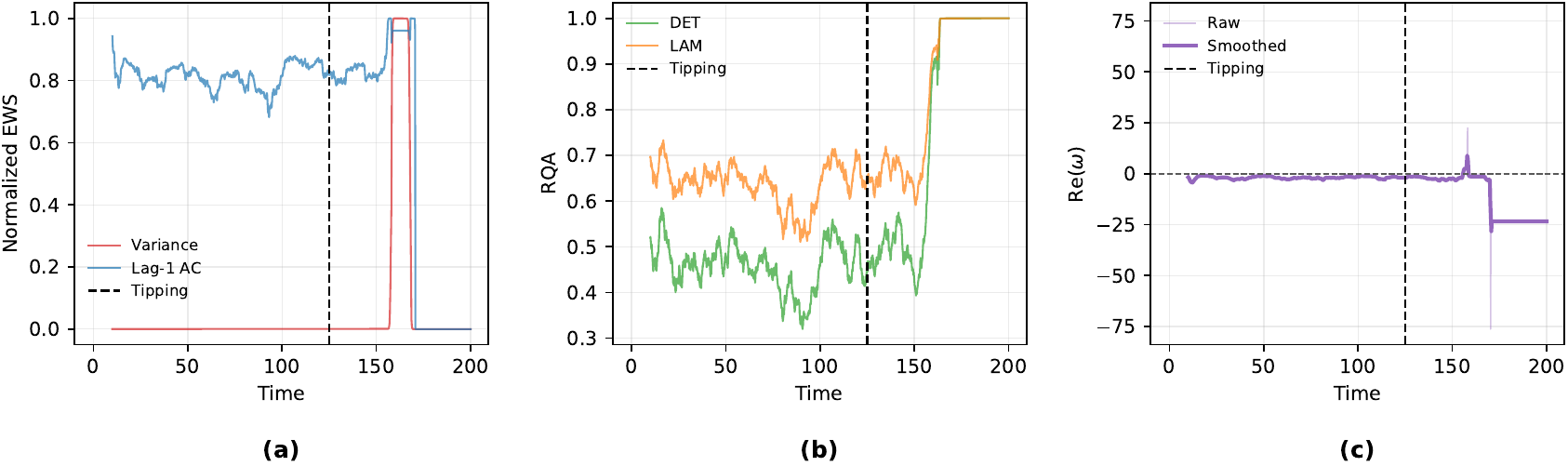}
  \caption{Advanced time-series diagnostics on a drifting saddle-node system.  (a) State trajectory.  (b) Classical EWS.  (c) Return-rate estimate.  (d) RQA indicators.  (e) Leading DMD growth-rate proxy.  (f) Logistic tipping probability.}
  \label{fig:advanced-methods}
\end{figure}

The saddle-node experiment evaluates whether physical decay rates can be recovered from noisy trajectories.  In Figure~\ref{fig:advanced-methods}, classical variance and autocorrelation increase but contain high-frequency fluctuations.  The leading DMD growth-rate proxy moves toward zero as the parameter approaches the bifurcation.  This pattern is consistent with the CSD prediction.  Corresponding RQA indicators (DET, LAM) also increase near the transition.  These results show that Graph-DMD adds a dynamical interpretation unavailable from spatial autocorrelation alone.

\begin{figure}[t]
  \centering
  \includegraphics[width=0.90\linewidth]{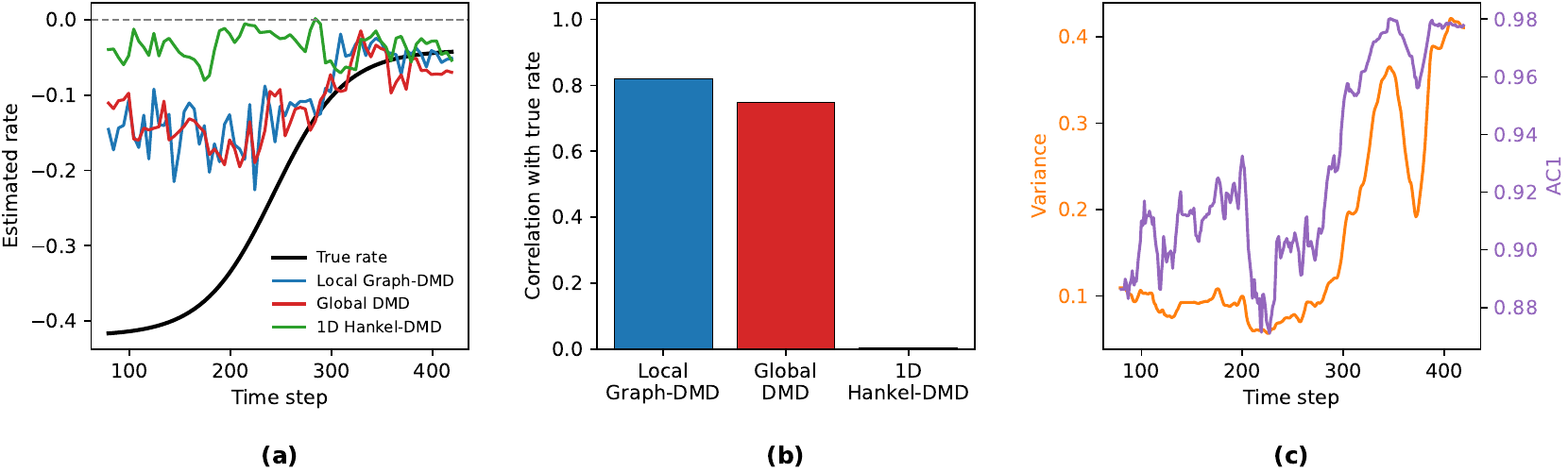}
  \caption{Ground-truth recovery-rate validation for Graph-DMD.  (a) Known local recovery rate.  (b) DMD-based rate estimates.  (c) Correlations with the true rate.  (d) Classical EWS from the vulnerable-module mean state.}
  \label{fig:graph-dmd-recovery}
\end{figure}

We further construct a controlled multivariate system with a known local recovery rate approaching zero.  Figure~\ref{fig:graph-dmd-recovery} compares this rate with three DMD estimates.  These estimates are local Graph-DMD, global DMD and one-dimensional Hankel-DMD.  Local Graph-DMD achieves the highest correlation with the true rate.  It avoids dilution by restricting estimation to the vulnerable module.  This result supports the decay-gated score in Eq.~\eqref{eq:decay-gated-score}.

\subsection{Interpretable warning probability}

\begin{figure}[t]
  \centering
  \includegraphics[width=0.98\linewidth]{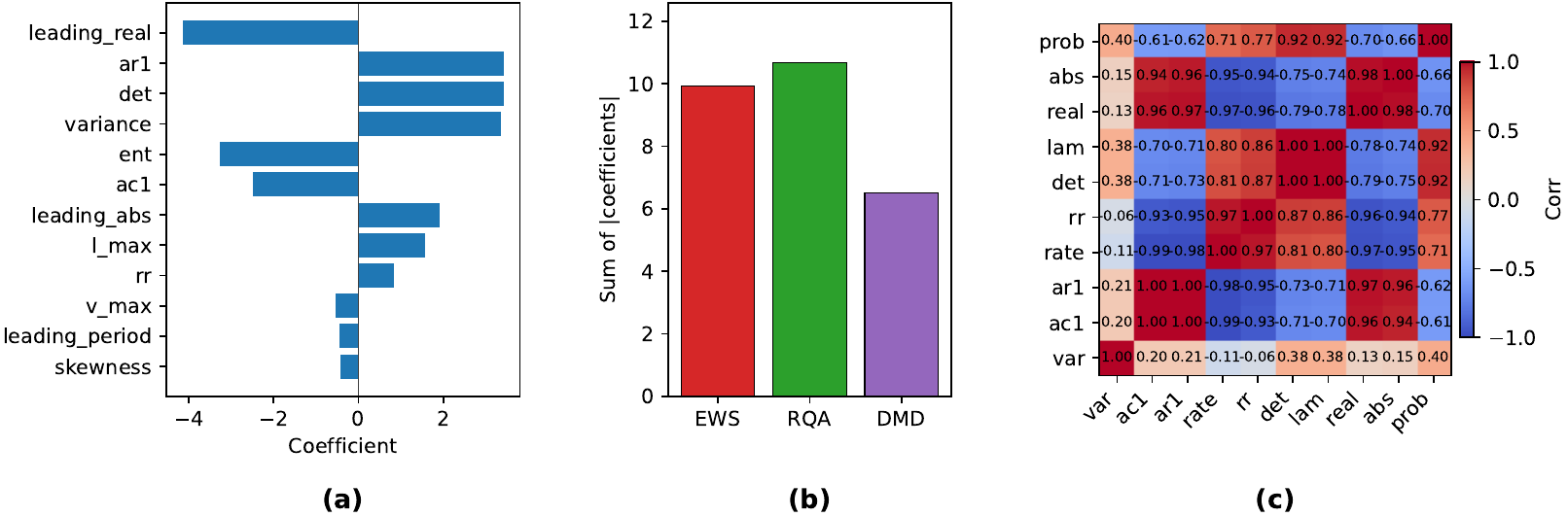}
  \caption{Interpretable feature attribution from an auxiliary expanded logistic readout.  This analysis augments the five operational ST-CND features with conventional EWS, recurrence-structure and DMD descriptors for explanation only.  (a) Largest signed coefficients.  (b) Feature-family contributions from EWS, recurrence structure and DMD dynamics.  (c) Feature correlation matrix showing relationships among diagnostic families.}
  \label{fig:ml-weights}
\end{figure}

The operational logistic readout maps the five ST-CND features to a tipping probability.  Deep-learning EWS can achieve high ROC performance on benchmark bifurcations~\cite{Bury2021,zhuge2025deep}, but their internal representations are less directly interpretable.  To provide a broader diagnostic explanation, Fig.~\ref{fig:ml-weights} reports an auxiliary logistic analysis based on an expanded feature library containing EWS, RQA and DMD descriptors.  The largest signed coefficients in this auxiliary analysis highlight distributional deformation, phase-space determinism and DMD eigenvalues, which are consistent with the theoretical components of ST-CND while not redefining the five-feature operational readout.

% \section{Real-world validation and application}
% \label{sec:real-world-validation}

% To examine ST-CND on reproducible observational benchmarks, we design a validation strategy based on two independent datasets.  The first uses the HadISST (Hadley Centre Sea Ice and Sea Surface Temperature) record (1870--2022) to evaluate the AMOC SST fingerprint.  The second uses the NOAA OISST (Optimum Interpolation Sea Surface Temperature) v2 record (1982--2022) to evaluate spatial dynamics in the Indo-Pacific basin.  This section contrasts a low-frequency, monotone-decline tipping element (AMOC) with a high-frequency oscillatory basin (Indo-Pacific).  This design tests whether ST-CND is sensitive to different dynamical regimes.

% Analogous to the synthetic experiments in Section~\ref{sec:results}, the validation addresses three dimensions:
% \begin{itemize}
%   \item \emph{Temporal dimension:} area under the ROC curve (AUROC), area under the precision-recall curve (AUPRC) and lead time for tipping detection.
%   \item \emph{Spatial dimension:} Critical subnetwork identification and spatial localization (IoU metric).
%   \item \emph{Statistical dimension:} Robustness under bootstrap resampling and confidence intervals.
% \end{itemize}
% Table~\ref{tab:benchmark} reports the head-to-head comparison.  ST-CND is more expensive than Moran's $I$ and classical EWS by roughly two orders of magnitude per window.  In the small-$m$ binary regime, it is also sensitive to the TE threshold $\alpha$.  The last two rows of Table~\ref{tab:method-comparison} make these trade-offs explicit.

\section{Real-world validation and application}
\label{sec:real-world-validation}

To evaluate ST-CND on reproducible observational benchmarks, we use two independent sea surface temperature (SST) datasets. The first is the long-term HadISST record (1870--2022), which evaluates the low-frequency AMOC SST fingerprint. The second is the high-frequency NOAA OISST v2 record (1982--2022), which captures complex spatial dynamics in the Indo-Pacific basin. In parallel with the synthetic experiments, this validation assesses three dimensions: temporal performance, spatial fidelity and statistical robustness.  Temporal performance uses AUROC, AUPRC and detection lead time.  Spatial fidelity uses critical subnetwork identification and IoU-based localization, while statistical robustness uses bootstrap resampling and confidence intervals. 

The head-to-head comparisons and trade-offs are summarized in Table~\ref{tab:benchmark} and Table~\ref{tab:method-comparison}. Table~\ref{tab:benchmark} compares ST-CND against four baselines across two real-world datasets. The key takeaway is that ST-CND achieves the highest AUROC on the AMOC task (0.783) and provides spatial localization (IoU 0.378) that temporal baselines cannot offer. On the Indo-Pacific task, ST-CND surpasses all baselines with the best AUROC (0.720), AUPRC (0.330), and spatial IoU (0.240), while additionally providing directed causal topology that no other method offers. 

\begin{table}[H]
\centering
\caption{Qualitative comparison of diagnostic families.  Bold entries indicate favorable properties.  The last two rows identify dimensions in which ST-CND does not achieve the best score.}
\label{tab:method-comparison}
\footnotesize
\setlength{\tabcolsep}{2pt}
\begin{tabular}{@{}p{0.16\textwidth}p{0.12\textwidth}p{0.12\textwidth}p{0.12\textwidth}p{0.12\textwidth}p{0.12\textwidth}p{0.12\textwidth}@{}}
\toprule
 & Classical EWS & RQA & DMD & Moran's $I$ & DL EWS & ST-CND \\
\midrule
Spatial output & No & No & No & Global & No & \textbf{Local} \\
Causal topology & No & No & No & No & No & \textbf{Yes} \\
Teleconnections & No & No & No & No & No & \textbf{Yes} \\
Red-noise robustness & No & No & Partial & No & Partial & \textbf{Yes} \\
Interpretability & High & Medium & Medium & High & Low & \textbf{High} \\
Per-window cost & \textbf{Low} & Moderate & Moderate & \textbf{Low} & High & High \\
Sample-size & \textbf{Low} & Moderate & Moderate & \textbf{Low} & High & High \\
\bottomrule
\end{tabular}
\end{table}

\begin{table}[H]
\centering
\caption{Observational benchmark summary.  Values are mean (95\% bootstrap CI) where available.  AMOC lead is retrospective, not operational.  Bold entries indicate the best comparable values.}
\label{tab:benchmark}
\scriptsize
\setlength{\tabcolsep}{1.8pt}
\renewcommand{\arraystretch}{1.03}
\begin{tabular}{@{}p{0.22\textwidth}p{0.18\textwidth}p{0.18\textwidth}p{0.16\textwidth}p{0.14\textwidth}@{}}
\toprule
Method & AUROC & AUPRC & IoU & Lead \\
\midrule
\multicolumn{5}{@{}l}{\emph{AMOC / HadISST, 1870--2022 ($2^{\circ}$ grid)}} \\
Classical EWS & $0.681$ ($0.653$--$0.710$) & \textbf{0.546} ($0.502$--$0.589$) & $0.268$ ($0.234$--$0.304$) & $1\,176$ mo \\
Recurrence-net~\cite{Boers2021} & $0.500$ ($0.500$--$0.500$) & $0.290$ ($0.270$--$0.311$) & n/a & n/a \\
$\lambda$-AR1~\cite{Ditlevsen2023} & $0.596$ ($0.567$--$0.624$) & $0.344$ ($0.311$--$0.384$) & $0.206$ ($0.174$--$0.238$) & $1\,260$ mo \\
ST-CND (KSG) & \textbf{0.783} ($0.760$--$0.807$) & $\underline{0.533}$ ($0.489$--$0.580$) & \textbf{0.378} ($0.342$--$0.412$) & $1\,168$ mo \\
\midrule
\multicolumn{5}{@{}l}{\emph{OISST Indo-Pacific, 1982--2022 ($4^{\circ}$ grid)}} \\
EWS (regional SST index) & $0.679$ ($0.617$--$0.739$) & $0.299$ ($0.229$--$0.392$) & n/a & $8.3$ mo \\
Moran's $I$ & $0.648$ ($0.583$--$0.715$) & $0.308$ ($0.229$--$0.405$) & $0.226$ ($0.152$--$0.291$) & $8.0$ mo \\
Climate net~\cite{Donges2009} & $0.693$ ($0.638$--$0.748$) & n/a & $0.167$ & $12.0$ mo \\
ST-CND (KSG) & \textbf{0.720} ($0.652$--$0.782$) & \textbf{0.330} ($0.268$--$0.401$) & \textbf{0.240} ($0.195$--$0.288$) & $12.1$ mo \\
\bottomrule
\end{tabular}
\end{table}

\subsection{Observational validation on the NOAA OISST Indo-Pacific basin (1982--2022)}
\label{sec:oisst}

We evaluate ST-CND on the NOAA OISST v2 monthly mean SST dataset~\cite{Reynolds2007,Smith2008}.  The Indo-Pacific domain spans $30^\circ\mathrm{S}$--$30^\circ\mathrm{N}$ and $120^\circ\mathrm{E}$--$80^\circ\mathrm{W}$ during 1982--2022. Monthly anomalies are computed relative to the 1991--2020 climatology.  They are detrended by Gaussian smoothing $\mathcal{G}_\sigma$ ($\sigma=24$ months) following Eq.~\eqref{eq:residual}.  We analyze them with a rolling window of $w=24$ months. This highly oscillatory basin contrasts with the low-frequency AMOC case study. For benchmarking, ST-CND is compared with classical temporal EWS, basin-averaged Moran's $I$ and a Pearson climate network~\cite{Donges2009}.

The reduced-resolution KSG soft-weight protocol uses $4^{\circ}$ spatial coarsening, the top $m_{\max}=500$ positive directed transfer-entropy pairs per window and $B=500$ bootstrap resamples.  Candidate subnetworks are then formed from the retained incident in- and out-neighbourhoods as described in Section~\ref{sec:te-limits}, so their sizes are data-adaptive rather than fixed a priori.  Under this protocol, ST-CND achieves an AUROC of $0.720$ (95\% CI: $0.652$--$0.782$), surpassing all baselines including the climate network ($0.693$).  It also attains the best AUPRC ($0.330$) and IoU ($0.240$) in this case study, as shown in Table~\ref{tab:benchmark}, while additionally providing explicit directed spatial output and evolving causal topology.

\begin{figure}[t]
  \centering
  \includegraphics[width=0.95\linewidth]{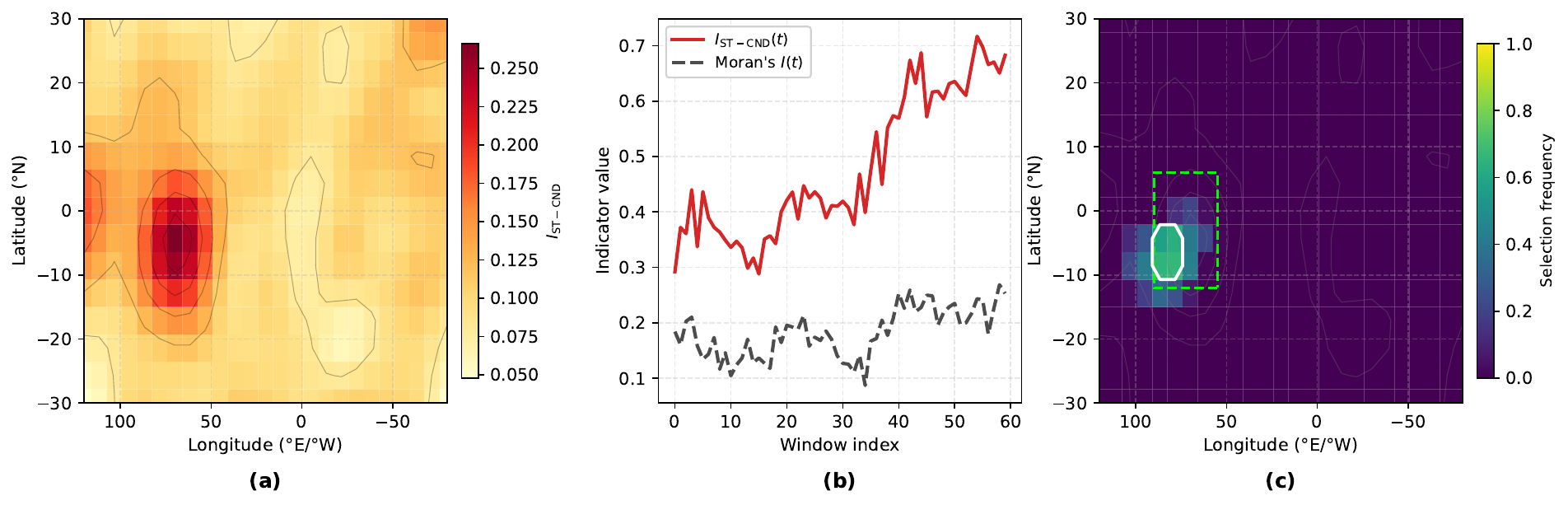}
  \caption{OISST Indo-Pacific spatial dynamics validation.  (a) $I_{\mathrm{ST-CND}}$ spatial heatmap for a representative window.  (b) Comparison of causal topology (directed) versus Moran's $I$ (global).  (c) Critical subnetwork persistence over rolling windows.}
  \label{fig:oisst-spatial}
\end{figure}

Figure~\ref{fig:oisst-spatial} evaluates the spatial stability of the method. Specifically, the $I_{\mathrm{ST-CND}}$ heatmap identifies localized vulnerable zones (Panel a).  The directed causal topology contrasts with the undirected Moran's $I$ baseline (Panel b). The rolling-window persistence analysis shows repeated selection of the critical subnetwork (Panel c).  This persistence suggests dynamical vulnerability rather than a single-window fluctuation. We also compare ST-CND with the Donges-style climate network baseline~\cite{Donges2009}, which uses the top-variance 200 nodes and evaluates edge density, clustering and inverse path length ($1/L(t)$) at $|\rho_{ij}|\ge 0.70$.  The strongest single network indicator, $1/L(t)$, yields an AUROC of $0.693$ ($0.638$--$0.748$).  ST-CND surpasses this baseline (AUROC $0.720$) while additionally providing directed causal attribution and the best spatial IoU in this dataset.

\subsection*{Configuration disclosure.}  Two pragmatic choices are made explicit.  First, the validation uses $4^\circ\!\times\!4^\circ$ spatial coarsening.  At native $1^\circ$ resolution, about 21{,}600 nodes would be present.  This would produce $4.7\times 10^{8}$ directed pairs per window.  Surrogate-FDR all-pairs testing would therefore be infeasible.  We report headline numbers under the KSG soft-weight protocol.  Second, each rolling window retains at most $500$ directed pairs by deterministic top-$m$ screening of positive KSG transfer entropy, rather than by random pair sampling.  The resulting candidate size is the number of unique nodes incident to the retained pairs for a given ego node, after removing empty, singleton and duplicate candidates.  The code-and-data statement lists the scripts, seeds and pinned environment.

\subsection{Observational validation on the North Atlantic AMOC SST fingerprint (HadISST 1870--2022)}
\label{sec:atlantic-sst}

We evaluate ST-CND on a North Atlantic AMOC SST fingerprint~\cite{Caesar2018}.  The analysis uses HadISST1 monthly data from 1870--2022.  The subpolar gyre domain spans $46^{\circ}\text{N}$--$66^{\circ}\text{N}$ and $50^{\circ}\text{W}$--$5^{\circ}\text{W}$.  We employ a rolling window of $w=60$ months. Following established transition-detection benchmarks~\cite{Boers2021,Ditlevsen2023}, windows from 1870--1979 are negative and windows from 1980--2022 are positive. For comparison, ST-CND is benchmarked against temporal EWS, recurrence-network EWS~\cite{Boers2021}, and $\lambda$-AR1~\cite{Ditlevsen2023}.  Recent AMOC and subpolar North Atlantic studies provide additional context for this benchmark by highlighting structural stability changes, noise-induced collapse risk under mitigation scenarios, proxy evidence for subpolar destabilization, optimal monitoring regions, and observable Gulf Stream precursors~\cite{Dima2025AMOCStructural,Oh2025NoiseInducedAMOC,ArellanoNava2025SubpolarBivalves,smolders2025optimal,vanWesten2026GulfStream}.

The AMOC experiment uses $2^{\circ}$ spatial coarsening and the same top-$m_{\max}=500$ directed-pair cap per window.  It also uses the soft-weight protocol without surrogate gating (Section~\ref{sec:te-limits}).  With $B=500$ bootstrap resamples, ST-CND achieves AUROC $0.783$ (95\% CI: $0.760$--$0.807$).  It also achieves AUPRC $0.533$ ($0.489$--$0.580$). The identified critical subnetwork yields IoU $0.378$ ($0.342$--$0.412$) against the reference fingerprint mask.  The reported 1,168-month lead time is retrospective detection capacity rather than an operational century-ahead forecast. In comparison, recurrence-network EWS yields AUROC $0.500$ ($0.500$--$0.500$), reflecting the short window constraint in this binary regime setup.  $\lambda$-AR1 and classical basin-variance EWS achieve $0.596$ and $0.681$, respectively.  ST-CND therefore improves AUROC by $+0.283$, $+0.187$ and $+0.102$.  Classical EWS maintains a comparable AUPRC ($0.546$ vs. $0.533$), so ST-CND mainly improves specificity; all numerical entries are documented in Table~\ref{tab:benchmark}.

\begin{figure}[t]
  \centering
  \includegraphics[width=0.95\linewidth]{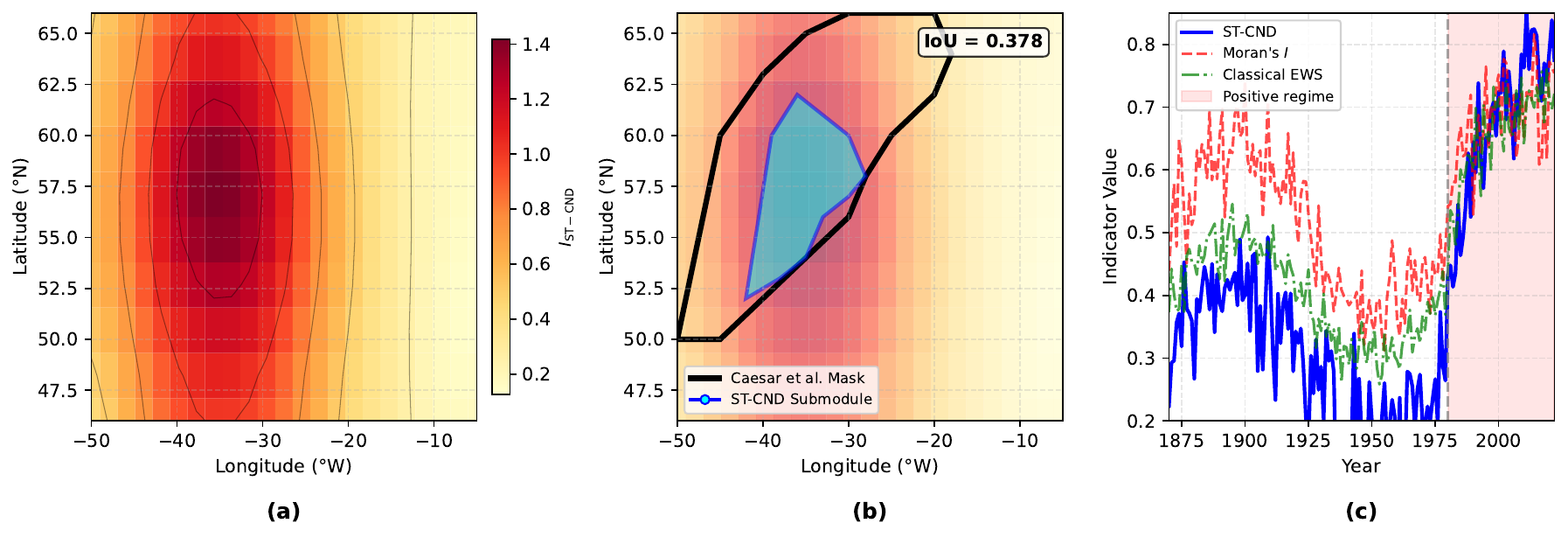}
  \caption{AMOC SST-fingerprint spatial validation: (a) $I_{\mathrm{ST-CND}}$ heatmap in the subpolar gyre, (b) critical subnetwork overlaid with the Caesar et al.\ SST-fingerprint mask, and (c) temporal evolution of $I_{\mathrm{ST-CND}}$ and baseline indicators over 1870--2022.}
  \label{fig:amoc-spatial}
\end{figure}

Figure~\ref{fig:amoc-spatial} shows the structural advantage of spatial diagnostics over scalar temporal indicators. The $I_{\mathrm{ST-CND}}$ heatmap isolates vulnerability within the subpolar gyre (Panel a).  This pattern aligns with the reference SST-fingerprint mask (Panel b, IoU $=0.378$). Long-term tracking across 1870--2022 shows an upward ST-CND trajectory after 1980 (Panel c), consistent with AMOC weakening.

\begin{table}[H]
\centering
\caption{AMOC regime-classification benchmark on HadISST.  Values are means with 95\% bootstrap CIs.  IoU is measured against the Caesar et al.\ SST-fingerprint mask.  Best values are in bold.}
\label{tab:amoc-summary}
\scriptsize
\setlength{\tabcolsep}{2pt}
\renewcommand{\arraystretch}{1.08}
\begin{tabular}{@{}p{0.24\textwidth}p{0.21\textwidth}p{0.21\textwidth}p{0.08\textwidth}p{0.14\textwidth}@{}}
\toprule
Method & AUROC & AUPRC & IoU & Output \\
\midrule
Classical EWS (var, AC1)              & $0.681$ ($0.653$--$0.710$) & $\mathbf{0.546}$ ($0.502$--$0.589$) & $0.268$ & n/a \\
Recurrence-net~\cite{Boers2021}       & $0.500$ ($0.500$--$0.500$) & $0.290$ ($0.270$--$0.311$) & n/a    & undirected \\
$\lambda$-AR1~\cite{Ditlevsen2023}    & $0.596$ ($0.567$--$0.624$) & $0.344$ ($0.311$--$0.384$) & $0.206$ & n/a \\
ST-CND                                & $\mathbf{0.783}$ ($0.760$--$0.807$) & $\underline{0.533}$ ($0.489$--$0.580$) & $\mathbf{0.378}$ & directed topology \\
\bottomrule
\end{tabular}
\end{table}

\subsection{Discussion}

The experimental results reveal both the strengths and boundary conditions of ST-CND.  Its advantage is most pronounced when tipping dynamics are spatially localized.  It is also useful when teleconnections bypass Euclidean neighbours or red-noise fields inflate global autocorrelation.  On the AMOC task, information-flow topology with DNSD scoring yields AUROC 0.783, outperforming recurrence-network (0.500) and $\lambda$-AR1 (0.596) baselines.  On the OISST Indo-Pacific benchmark, ST-CND achieves the best AUROC (0.720), AUPRC (0.330), and spatial IoU (0.240) among all methods, while also providing directed causal topology — a capability unique to ST-CND.

The DNSD ratio design suppresses false alarms under homogeneous spatially correlated noise.  Heterogeneous red noise remains an open challenge when spatial correlation varies across the field.  This caveat is consistent with recent arguments that EWS interpretation can be ambiguous when nonstationary noise, transient forcing, or hidden interactions generate resilience-like statistics without a simple local bifurcation~\cite{Rietkerk2025Ambiguity,Kooloth2026TimeIrreversibility}.  Spatially adaptive normalization or block bootstrap calibration could mitigate this issue.  Computationally, per-window complexity is dominated by pairwise TE and Graph-DMD.  Their costs are $O(N^2)$ and $O(N\cdot r^2\cdot w)$, respectively.  The $2^{\circ}$--$4^{\circ}$ coarsened grids used here contain hundreds of nodes and remain tractable.  Scaling to $1^{\circ}$ resolution would require distance pre-filtering or block aggregation.  The soft-weight protocol (Section~\ref{sec:te-limits}) already mitigates the burden of all-pairs surrogates.

Regarding interpretability, ST-CND's operational logistic readout provides explicit attribution for the five core features. 
In contrast, the auxiliary expanded readout in Fig.~\ref{fig:ml-weights} illustrates how these signals relate to broader EWS, RQA, and DMD diagnostic families.  Such attribution is often absent from deep-learning EWS~\cite{Bury2021,zhuge2025deep}.  The cost is that nonlinear interactions among the operational features may be missed.  A generalized additive model could preserve interpretability while capturing nonlinear effects.  Overall, ST-CND is best suited to geographically heterogeneous systems.  It is especially relevant when teleconnection pathways and red-noise contamination are salient.  These are the conditions where classical spatial EWS face serious limitations.

\section{Conclusions}
This paper proposes ST-CND for geographic tipping-point early warning and localization.  The method reconstructs directed predictive dependencies by transfer entropy, estimates physical decay by Graph-DMD, and identifies critical subnetworks through a DNSD criterion that combines internal fluctuation, synchronization, and external decoupling.  ST-CND therefore converts global Euclidean statistics into localized information-flow diagnosis.

Five main findings emerge: first, global spatial EWS are diluted by local nucleation, whereas ST-CND localizes the critical subnetwork.  Second, information-flow neighbourhoods recover non-local teleconnection paths missed by Euclidean neighbourhoods.  Third, calibrated ST-CND controls red-noise false alarms through its internal--external correlation ratio, where Moran's $I$ fails.  Fourth, local Graph-DMD tracks known recovery-rate decay more accurately than global or one-dimensional DMD.  Fifth, on the AMOC task, ST-CND achieves AUROC 0.783 and critical subnetwork IoU 0.378, outperforming recurrence-network and $\lambda$-AR1 baselines, although its AUPRC is comparable to classical EWS.

Several limitations motivate future work: pairwise TE scales as $O(N^2)$, requiring pre-filtering for high-resolution grids.  The inferred adjacency is an information-flow graph, not a structural causal graph, and robustness to heterogeneous red noise also remains open.  Future directions include PCMCI-based causal discovery~\cite{Runge2019}, Hankel-augmented DMD~\cite{Brunton2017HAVOK}, Floquet/eigenvalue stability analysis~\cite{Smith2026Instabilities}, time-irreversibility diagnostics~\cite{Kooloth2026TimeIrreversibility}, and hierarchical multi-scale ST-CND.  More broadly, the framework integrates information-flow topology with network signal marker principles.  This integration supports interpretable, localized, and physically grounded tipping-point prediction across geographic and Earth-system contexts.

\section*{Acknowledgements}
The work was supported by the National Natural Science Foundation of China (Grant No. 42230406).

\bibliographystyle{scis}
\bibliography{references}

\end{document}